%% file: paper.tex
\documentclass[conference]{IEEEtran}
\IEEEoverridecommandlockouts
\usepackage{cite}
\usepackage{amsmath,amssymb,amsfonts}
\usepackage{algorithmic}
\usepackage{graphicx}
\usepackage{textcomp}
\usepackage{xcolor}
\usepackage{subfig}

\usepackage{algorithm}

\def\BibTeX{{\rm B\kern-.05em{\sc i\kern-.025em b}\kern-.08em
    T\kern-.1667em\lower.7ex\hbox{E}\kern-.125emX}}
\begin{document}

\title{Autonomous Driving with a Deep Dual-Model Solution for Steering and Braking Control\\
\thanks{This work has been supported in part by the European Union’s Horizon
Europe research and innovation programme under grant agreement
No 101079214 (AIoTwin) and by Croatian Science Foundation under the project IP-2019-04-1986 (IoT4us).}
}

\author{\IEEEauthorblockN{Ana Petra Jukić, Ana Šelek, Marija Seder, Ivana Podnar Žarko}
\IEEEauthorblockA{
\textit{University of Zagreb Faculty of Electrical Engineering and Computing}\\
Zagreb, Croatia \\
ana-petra.jukic@fer.hr, ana.selek@fer.hr, marija.seder@fer.hr, ivana.podnar@fer.hr}
}

\maketitle

\begin{abstract}
The technology of autonomous driving is currently attracting a great deal of interest in both research and industry. In this paper, we present a deep learning dual-model solution that uses two deep neural networks for combined braking and steering in autonomous vehicles. Steering control is achieved by applying the NVIDIA's PilotNet model to predict the steering wheel angle, while braking control relies on the use of MobileNet SSD. Both models rely on a single front-facing camera for image input. The MobileNet SSD model is suitable for devices with constrained resources, whereas PilotNet struggles to operate efficiently on smaller devices with limited resources. To make it suitable for such devices, we modified the PilotNet model using our own original network design and reduced the number of model parameters and its memory footprint by approximately 60\%. The inference latency has also been reduced, making the model more suitable to operate on resource-constrained devices. The modified PilotNet model achieves similar loss and accuracy compared to the original PilotNet model. When evaluated in a simulated environment, both autonomous driving systems, one using the modified PilotNet model and the other using the original PilotNet model for steering, show similar levels of autonomous driving performance.
\end{abstract}


\input{section/1_introduction.tex}
\input{section/2_related_work.tex}

\input{section/3_arch}

\input{section/4_results}

\input{section/5_conclusion}

\bibliographystyle{IEEEtran}
\bibliography{IEEEabrv,mybibfile}

\end{document}

%% file: section/1_introduction.tex
\section{Introduction}

Recent advances in computing power and deep learning have revolutionized autonomous vehicles. Deep learning enables accurate perception of the environment through feature extraction from sensor data, leading to improved decision-making capabilities. Its ability to learn from large datasets enables the design of autonomous vehicles that can effectively handle various driving scenarios.\par

Traditional autonomous driving systems consist of perception, localization, spatial mapping, path planning, and trajectory tracking control, where each step is computationally demanding, especially in complex environments with a large number of obstacles \cite{autonomous_driving_handbook}. Traditional systems typically rely on finely defined rules and complex algorithms \cite{autonomous_vehicles_review}, which engineers have to create manually. Modern autonomous driving systems are based on deep learning methods and use neural networks to control the driving scenarios of autonomous vehicles. Instead of manually defined rules and algorithms, deep learning solutions use large datasets to learn how to make decisions autonomously while driving. Deep learning allows the system to extract important driving information from raw data, unlike traditional systems, which require detailed data analysis. The traditional approach may be more robust and easier to understand and interpret, but it requires a large amount of manual algorithm definition. On the other hand, deep learning enables automated learning of important features from data, decreasing the amount of manually programmed code~\cite{deep_learning_vehicle_control}. However, deep learning models can be complex to interpret and require a large amount of computational resources and data. In recent years, a combination of traditional approaches and deep learning is often used to leverage the strengths of both approaches and develop advanced autonomous driving systems. 

Deep learning methods are utilized for various autonomous driving tasks, including scene classification, driving control, scene understanding, path planning, pedestrian and obstacle detection, lane tracking, and recognition of traffic lights and signs. State of the art solutions leverage convolutional networks, recurrent neural networks, auto-encoders and deep reinforcement learning for tasks such as classification, regression, recognition, detection, segmentation and prediction \cite{deep_learning_survey}. \par

End-to-end learning for autonomous vehicles is a deep learning approach in which a single neural network learns to map raw sensor inputs, e.g. from cameras, radar and/or lidar, to vehicle controls such as acceleration and throttle. In contrast to the traditional approach, where each step typically involves a specific algorithm, the end-to-end model optimizes multiple tasks by training a single neural network. Three main methodologies for end-to-end learning include: supervised deep learning where a neural network is trained on pairs of input data (such as images from cameras) and corresponding target outputs (such as steering angles or throttle/brake commands), deep reinforcement learning in which a self driving car learns through trial and error interactions with an environment, and neuroevolution where evolutionary algorithms are used for training neural networks \cite{autonomous_driving_survey}. 
End-to-end driving shows potential, but it still has not been fully used in real city situations, except for a few demonstrations. The main problems of end-to-end driving are the lack of interpretability and inability to predefine safety rules to prevent accidents and ensure safe driving. \par

The PilotNet end-to-end model is a convolutional neural network that maps raw images captured by front-facing cameras directly to steering angle of the steering wheel. The model was first described in \cite{pilotnet}, where the authors trained the model to drive on roads with or without lane markings in various weather conditions and during day and night. The main aim was to ensure an autonomous vehicle's ability to stay on a road and maintain its lane. The model learned to identify important features corresponding to the recognition of road and lane markings based only on the image of the road and the steering wheel angle at that moment. The evaluation of the model was conducted by comparing the duration of autonomous driving with the duration of instances requiring human intervention when the vehicle deviated more than one meter from the center line. Each instance of human intervention was estimated to last 6 seconds. The model was evaluated on real roads, where a vehicle traveled approximately 20 km with the vehicle operating autonomously for 98\% of the time. Not a single human intervention was recorded over a distance of 16 km. Originally, PilotNet is designed to run on NVIDIA DRIVE\textsuperscript{TM} PX, but it has difficulty running on devices with limited computational resources, such as the Raspberry Pi Pico \cite{deeppicarmicro}. \par
 
The autonomous driving system with a deep dual-model solution designed and evaluated in this paper effectively navigates the road while maintaining its lane and incorporates a mechanism to prevent collisions with vehicles ahead. Input images come exclusively from a single front-facing camera. Collision avoidance works by slowing down the autonomous vehicle when it approaches too close to a vehicle ahead. The system employs two neural networks: PilotNet, which determines the steering wheel angle, and MobileNet SSD, which detects vehicles on the road and estimates their distance from the autonomous vehicle. The MobileNet SSD model is a real-time detection model that identifies the object's class and its bounding box in an the input image. MobileNet handles image classification while SSD (Single Shot MultiBox Detector) is responsible for the detection of objects. Both MobileNet and SSD are convolutional neural networks, with MobileNet utilizing depthwise separable convolution for  reasons. This model is suitable for use on mobile and embedded devices with limited resources. 

The use of two separate models for steering and braking in an autonomous driving system offers several advantages compared to a system utilizing only one model (end-to-end model) for both tasks. It enables specialization, so that each model is optimized for its particular task. Additionally, it enhances safety by allowing one model to operate independently even if the other fails, which reduces the risk of accidents. Moreover this approach allows for easier customization, debugging and modification of components within the system. 

Although an autonomous driving system such as this can operate smoothly in real time on devices with sufficient computational power, those with limited computational capacity can encounter inference delays. To address this issue, we modified the original PilotNet model by using depthwise separable convolution and bottleneck layers to reduce the number of model parameters and inference latency, while maintaining a comparable loss to the original PilotNet model. The new PilotNet model was trained using the data gathered from driving simulations in Grand Theft Auto V (GTA V), while we utilized a pretrained MobileNet SSD model trained with the COCO dataset. The autonomous driving system was evaluated within GTA V in different weather and light conditions and road types.

Our contribution can be summarized as follows:
\begin{itemize}
  \item We developed an autonomous driving system with a deep dual-model solution capable of following the road and avoiding crashes with the vehicles ahead, by using the PilotNet and MobileNet SSD models and a single front-facing camera as sensory input.
  \item We enhanced the PilotNet model to decrease the number of parameters, model size and inference latency, so that it can be used on resource-constrained devices.
  \item We developed controls for the autonomous driving system within GTA V and evaluated the performance of the deep dual-model across various scene conditions.
\end{itemize}

The paper is organized as follows. Section \ref{related_work} provides an overview of prior research in the field of autonomous driving solutions developed and tested using GTA V, as well as modifications made to the PilotNet model in prior research. Section \ref{arch} provides an overview of the proposed autonomous driving system, the setup in GTA V to simulate autonomous driving and data collection, alterations to the PilotNet, the training process for both the original and modified PilotNet, and their comparison. Section \ref{results} provides the results of simulation used to evaluate the developed autonomous driving system in GTA V. Lastly, section \ref{conclusion} presents the conclusion and outlines potential areas for future research.

%% file: section/2_related_work.tex
\section{Related work} \label{related_work}

Bechtel et al. \cite{deeppicarmicro} developed DeepPicarMicro, an autonomous RC car platform that runs PilotNet model for end-to-end steering on Raspberry Pi Pico, in a real-world environment. Due to its constrained computing resources, executing a deep learning model on a Raspberry Pi poses challenges in processing data within the required time constraints. Initially, PilotNet was trained, but its time-related performance suggested it would not effectively control the car. This is because PilotNet requires 3 seconds to process each frame, whereas the car's control period is only 133 milliseconds. To fix this issue, standard convolutional layers in PilotNet were replaced with depthwise separable layers, while retaining the same hyperparameters. However, this modification still resulted in unsatisfactory performance, with frame processing taking over 500 miliseconds. To improve efficiency, a network architecture search (NAS) was conducted on PilotNet with depthwise separable convolutions, with various network depths and layer widths. The performance of each model was then evaluated in a real-world scenario. It was observed that model accuracy alone did not adequately predict true model performance on the track, rather that inference latency also had significant impact. Therefore, the authors developed a joint optimization strategy that considered both accuracy and latency. This approach generally proved to be effective in accurately predicting each model's real-world performance. \par

Novello et al. \cite{gta_vgg_lstm} developed an end-to-end model that translates car hood camera image into a sequence of three driving commands: steering wheel angle, accelerator pedal pressure and brake pedal pressure in GTA V. This model is composed with both convolutional and recurrent neural network. The convolutional part is composed of the pretrained VGG16 network, while both components of the model are augmented with dense layers and their outputs are concatenated and further processed by dense layers for calculating the final model output. Data collection and evaluation were conducted in various weather conditions, with no other vehicles present on the road. Driving controls are implemented using the vJoy driver and x360ce software. Evaluation shows that an autonomous vehicle is able to drive similarly to a human driver. \par

Jaladi et al. \cite{gta_vgg_pilotnet} developed an end-to-end model suitable for driving on the highway in GTA V. The model predicts steering and throttle values to control the vehicle based on the screen image captured from the game. The two main objectives are training the vehicle to follow the road and to avoid collisions with vehicles ahead. Two distinct models were trained: VGG19 and PilotNet. In the VGG19 model, the weights for the convolution layers are pretrained and fixed, while only the weights of the fully connected layers are trained. The controls are implemented using the vJoy and an Xbox controller. The VGG19 model was pretrained with the ImageNet dataset, leading to faster convergence compared to Nvidia's architecture, which was trained from scratch. During evaluation, the VGG19 model consistently outperformed PilotNet. \par

Martinez et al. \cite{gta_alexnet} collected more than 480,000 labeled images depicting typical highway driving scenarios in GTA V. These images were used to train AlexNet to predict several scenario aspects including distance to lane markings, vehicles ahead and steering angle. The authors derived 8 affordance variables: steering angle, distances to neighboring cars in the left, center, and right lanes, distances to lane markings on both sides that are close to the vehicle and distances to lane markings on both sides that are further from the vehicle. The model was designed to extract 8 different affordance variables from each image using modifications within GTA V. Data collection was conducted across various road types, day and night conditions, and different weather conditions. The model showed good performance in estimating steering angle and distance to lane markings, but its accuracy was comparatively lower when estimating distances to other vehicles. The authors also highlighted several limitations of using GTA V as an autonomous driving simulator: the game was not intended as an academic tool and using it as such can lead to legal implications, data collection and evaluation requires game modifications, the game does not inherently support add-ons or custom 3D models and there are limitations in simulating smaller realistic details such as motion blur.  \par

%% file: section/3_arch.tex
\section{Autonomous Driving Using a Deep Dual-Model} \label{arch}
\begin{figure}[htbp]
\centerline{\includegraphics[width=0.9\columnwidth]{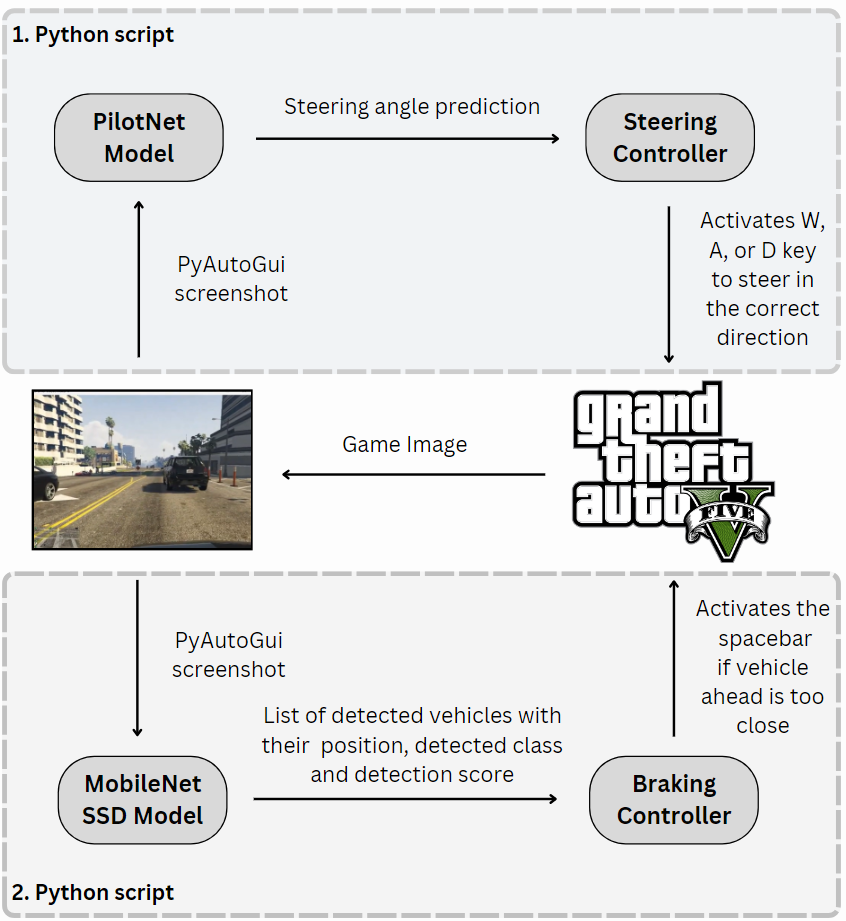}}
\caption{Architecture of the proposed autonomous driving system}
\label{fig:system_overview}
\end{figure}

The developed autonomous driving system employs two neural networks: PilotNet to determine the steering wheel angle, and MobileNet SSD to detect vehicles on the road and estimate their distance from the autonomous vehicle. Fig.~\ref{fig:system_overview} depicts the architecture of the implemented autonomous driving system which is divided into three components that function simultaneously: two Python processes, the first one for PilotNet and steering and the second one for MobileNet SSD and breaking, as well as the GTA V window.

The first Python process captures a screenshot of the GTA V window that displays the view of the front-facing camera mounted on the hood of an autonomous vehicle. The captured screenshot is forwarded as input to the PilotNet model, which predicts the steering wheel angle. The steering controller, within the Python process, simulates driving in GTA V by pressing the 'W', 'A' and 'D' keys on the keyboard using the Keyboard Python module. It determines which key to press based on the steering angle. As the steering angle increases or decreases, so does the sharpness of the turn that the vehicle is approaching. This situation can become problematic if the vehicle enters sharp turns at high speeds, potentially leading it to lose control and swerve off the road. To prevent this, the vehicle's speed is regulated by pressing the 'S' key and extending the process sleep time according to the steering angle, which decreases its speed in accordance with the sharpness of the turn.
The second Python process also captures a screenshot of the GTA V window, which is then used as input into the MobileNet SSD model to detect vehicles in front of the self-driving vehicle. The model outputs a list of detected vehicles along with the coordinates of the bounding box surrounding them and the probability of belonging to the vehicle class.
The coordinates of the bounding box are used to determine whether the detected vehicles are in the same lane as the autonomous vehicle and whether they are positioned in front of it. The proximity of a vehicle to the autonomous vehicle aligns with the probability that the detected object belongs to the vehicle class, considering that for larger objects (objects that take up more pixels in the input image), the model is more confident in its detection, therefore increasing the probability. When the nearest detected vehicle (the one with the highest probability of belonging to the vehicle class) surpasses a certain threshold probability, it is considered too close and the braking controller activates the spacebar to prevent collision.

\subsection{Simulator Preparation}
Modifications were added to GTA V to prepare it for data collection and self-driving simulation. Script Hook V with Native Trainer \footnote{ScriptHookV modification downloaded from https://www.gta5-mods.com/tools/script-hook-v.} was utilized for manipulating the game's world and implementing a speed limiter. The speed limiter was used due to the absence of throttle prediction in the self-driving car system. Its purpose was to ensure that the vehicle maintained a safe speed, especially when entering sharp turns, preventing it from potentially losing control and swerving off the road. 
Additionally, the Hood Camera mod \footnote{Hood camera modification downloaded from https://www.gta5-mods.com/scripts/hood-camera.} was used to incorporate a front-view camera for the vehicle. 

To capture the steering angle of the self-driving car at a specific moment, we utilize a virtual joystick implemented through the vJoy driver \footnote{VJoy driver downloaded from https://vjoy.software.informer.com/.} which is connected to the x360ce software \footnote{X360ce software downloaded from https://vjoy.software.informer.com/.}. 

\subsection{Data Collection}
The data consists of images captured by the vehicle's front camera along with the corresponding steering angle of the steering wheel at that particular moment. The data is collected by driving around the GTA V map using the keyboard to control the vehicle. X360ce software is responsible for translating keyboard inputs of the human driver into respective steering angles. The translated data is then gathered through the Pvjoy library, while simultaneously capturing images of the road using the PyAutoGui library in Python. Data was collected during driving sessions in both sunny and rainy weather, covering highway and city roads, and including both day and night conditions. 

The initial 100,000 data samples were captured, maintaining a frame rate limit of 10 fps. The image dimensions were 160x120x1 pixels, and the steering angles ranged from -1 (representing a full left turn) to 1 (representing a full right turn), with 0 indicating driving straight ahead. 
Since the majority of collected data primarily consisted of instances where the driver drove straight forward, dataset balancing was necessary. Consequently, redundant data points related to driving forward were removed, resulting in around 40,000 remaining data samples. Each data instance in the remaining dataset underwent image flipping, and its corresponding steering angle was multiplied by -1. This data was then added to the dataset, resulting in a total of 80,000 samples used for training and evaluating the models.

\subsection{Network Architectures and Training}
We designed a neural network inspired by the architecture of the PilotNet model, which utilizes depthwise separable convolution to reduce the number or parameters. Table \ref{tab1} shows the PilotNet network architecture, which consists of a normalization layer, five convolutional layers and four fully connected layers. 

\begin{table}[htbp]
\caption{PilotNet model architecture}
\begin{center}
    \centering
    \begin{tabular}{|c|c|c|c|c|}
        \hline
        \textbf{Layer} & \textbf{Kernel} & \textbf{Stride} & \multicolumn{1}{c|}{\textbf{Output}} & \textbf{Parameters}\\
        & & & \textbf{Channels} & \\
        \hline
        Normalization & - & - & 1 & 0\\
        \hline
        Conv2D & 5x5 & 2x2 & 24 & 624\\
        \hline
        Conv2D & 5x5 & 2x2 & 36 & 21636\\
        \hline
        Conv2D & 5x5 & 2x2 & 48 & 43248\\
        \hline
        Conv2D & 3x3 & 1x1 & 64 & 27712\\
        \hline
        Conv2D & 3x3 & 1x1 & 64 & 36928\\
        \hline
        Flatten & - & - & 2880 & 0\\
        \hline
        Dense & - & - & 100 & 665700\\
        \hline
        Dense & - & - & 50 & 5050\\
        \hline
        Dense & - & - & 10 & 510\\
        \hline
        Dense & - & - & 1 & 11\\
        \hline
        \multicolumn{4}{|c|}{} & Total: 801,419\\
        \hline
    \end{tabular}
    \label{tab:my_table}
\label{tab1}
\end{center}
\end{table}

Table \ref{tab1} depicts the network architecture of the modified PilotNet model. We replaced all convolutional layers in the original model with depthwise separable layers that maintain the same number of filters, except for the final depthwise separable layer where the number of filters is adjusted to 36, to further reduce the number of parameters. Depthwise separable convolution achieves parameter reduction and computational efficiency by breaking down the convolution operation into two separate steps and by working with smaller tensors. This improved computational efficiency consequently and reduced inference latency. Two bottleneck layers are inserted between the first and second, as well as between the third and fourth depthwise separable convolutional layers. These bottleneck layers consist of 1x1 standard convolutional layers, placed before the computationally intensive depthwise separable convolutions. As a result, this speeds up both training and inference. Bottleneck layers also help to improve model generalization by encouraging the network to learn more concise data representations. Both original PilotNet and its modified version employ ReLu activation function within their convolutional and dense layers. The original PilotNet model contains a total of 801,419 learnable parameters, whereas the modified PilotNet model contains a total of 303,180 learnable parameters, resulting in a reduction of approximately 62\%. The total count of learnable parameters is determined based on the size of the input image.

\begin{table}[htbp]
\caption{Modified PilotNet model architecture}
\begin{center}
    \centering
    \begin{tabular}{|c|c|c|c|c|}
        \hline
        \textbf{Layer} & \textbf{Kernel} & \textbf{Stride} & \multicolumn{1}{c|}{\textbf{Output}} & \textbf{Parameters}\\
        & & & \textbf{Channels} & \\
        \hline
        SeparableConv2D & 5x5 & 2x2 & 24 & 73\\
        \hline
        Conv2D & 1x1 & 1x1 & 12 & 300\\
        \hline
        SeparableConv2D & 5x5 & 2x2 & 48 & 924\\
        \hline
        SeparableConv2D & 5x5 & 2x2 & 36 & 2964\\
        \hline
        Conv2D & 1x1 & 1x1 & 18 & 666\\
        \hline
        SeparableConv2D & 5x5 & 2x2 & 64 & 1666\\
        \hline
        SeparableConv2D & 3x3 & 1x1 & 36 & 2916\\
        \hline
        Flatten & - & - & 2880 & 0\\
        \hline
        Dense & - & - & 100 & 288100\\
        \hline
        Dense & - & - & 50 & 5050\\
        \hline
        Dense & - & - & 10 & 510\\
        \hline
        Dense & - & - & 1 & 11\\
        \hline
        \multicolumn{4}{|c|}{} & Total: 303,180\\
        \hline 
    \end{tabular}
    \label{tab:my_table}
\label{tab2}
\end{center}
\end{table}

The original PilotNet and its modified version were implemented using TensorFlow in Python. The training set consisted of 65,000 samples from the dataset, while the remaining 15,000 were reserved for evaluation. Each model was trained with a batch size of 300 using the Adam optimizer with the mean square error (MSE) loss function. A batch generator was utilized during training, randomly selecting 300 samples at each step. Additionally, each selected sample underwent image augmentation with a 50\% chance, involving zooming, horizontal flipping, or brightness adjustment, with the goal of enhancing the dataset. The training process utilized an Nvidia GeForce RTX 3070 Ti GPU. The original PilotNet was trained for approximately 1000 epochs, while the modified version underwent training for approximately 2000 epochs to achieve loss comparable  to the original model. The test loss for the original PilotNet was 0.0205, with a mean absolute error (MAE) on the test set of 0.1035. The modified PilotNet exhibited a test loss of 0.0196, with a corresponding test set MAE of 0.1004. \par

Table \ref{tab3} shows the difference in performance and size between the modified PilotNet and the original PilotNet model. When saved on a computer, the modified PilotNet occupies 3.66 MB of memory, whereas the original PilotNet model occupies 9.48 MB. This indicates a reduction in size of approximately 61\%, corresponding to the decrease in the number of parameters. When comparing the performance of prediction for one frame, both RAM usage and inference latency are reduced. The inference latency, measured while the models utilize the 12th Gen Intel(R) Core(TM) i7-12700H CPU, decreases from roughly 246 ms to about 167 ms.

\begin{table}[htbp]
\caption{Comparison between the original and modified PilotNet model.}
\begin{center}
    \centering
    \begin{tabular}{|c|c|c|c|c|c|}
        \hline
        \textbf{Model} & \textbf{Inference Latency} 
        & \textbf{Size} & \textbf{RAM usage}\\
        \hline
         PilotNet & $\sim$246 ms & 9.48 MB & $\sim$367 MB\\
        \hline
         Modified PilotNet & $\sim$167 ms & 3.66 MB & $\sim$314 MB\\
        \hline
    \end{tabular}
    \label{tab:my_table}
\label{tab3}
\end{center}
\end{table}

%% file: section/4_results.tex
\section{Experimental results} \label{results}
Autonomous driving
solutions are evaluated in GTA V, with one system utilizing the original PilotNet and the other 
employing the modified model version. Systems are evaluated on both highway and city roads during the day and night, and in different weather conditions including rain and sunny weather. The PilotNet models were executed on 12th Gen Intel(R) Core(TM) i7-12700H CPU, while MobileNet SSD was run on the Nvidia GeForce RTX 3070 Ti GPU. Results are presented in \ref{tab4}. Each driving session lasted  five minutes. Following the methodology used in \cite{pilotnet}, the driving autonomy percentage was calculated using the frequency of human interventions. These interventions typically lasted for 6 seconds, which is the time required to realign the vehicle within the correct lane by pressing the corresponding keys on the keyboard. Both autonomous driving systems exhibit similar levels of autonomy. Small differences in autonomy arise due to the inability of replicating identical scenarios repeatedly in GTA V. Driving on the highway exhibited a higher percentage of autonomous driving compared to city road driving, as there is less traffic on the highway and fewer intersections compared to city roads. Daytime driving showed better results than nighttime driving. Driving in sunny weather yielded the best results, while driving in rain resulted in the worst performance. 

\begin{table}[htbp]
\caption{Evaluation of autonomous driving systems in GTA V, with the original and modified PilotNet model.}
\begin{center}
    \centering
    \begin{tabular}{|c|c|c|c|c|}
        \hline
        \shortstack{\textbf{Road Type} \\ \empty \\ \empty} & \shortstack{\textbf{Time} \\ \empty \\ \empty} & \shortstack{\textbf{Weather} \\ \empty \\ \empty} &
        \multicolumn{2}{c|}{\shortstack{\textbf{Number of Interventions/}\\ \textbf{Percentage of Autonomy}} } \\
        \cline{4-5}
        & & & \multicolumn{1}{c|}{PilotNet} & \multicolumn{1}{c|}{\shortstack{Modified \\ PilotNet}}\\
        \hline
        Highway & Day & Sunny & 0, 100\% & 0, 100\%\\
        \hline
        Highway & Day & Rain & 7, 86\% & 6, 88\%\\
        \hline
        Highway & Night & Clear sky & 4, 92\% & 3, 94\%\\
        \hline
        Highway & Day & Rain & 11, 78\% & 12, 76\%\\
        \hline
        City road & Day & Sunny & 5, 90\% & 4, 92\%\\
        \hline
        City road & Day & Rain & 12, 76\% & 14, 72\%\\
        \hline
        City road & Night & Clear sky & 7, 86\% & 7, 86\%\\
        \hline
        City road & Day & Rain & 17, 66\% & 20, 60\%\\
        \hline
    \end{tabular}
    \label{tab:my_table}
\label{tab4}
\end{center}
\end{table}
\label{solution}
\label{results}

\begin{figure}[htbp]
\centerline{\includegraphics[width=3in, height = 3.5in]{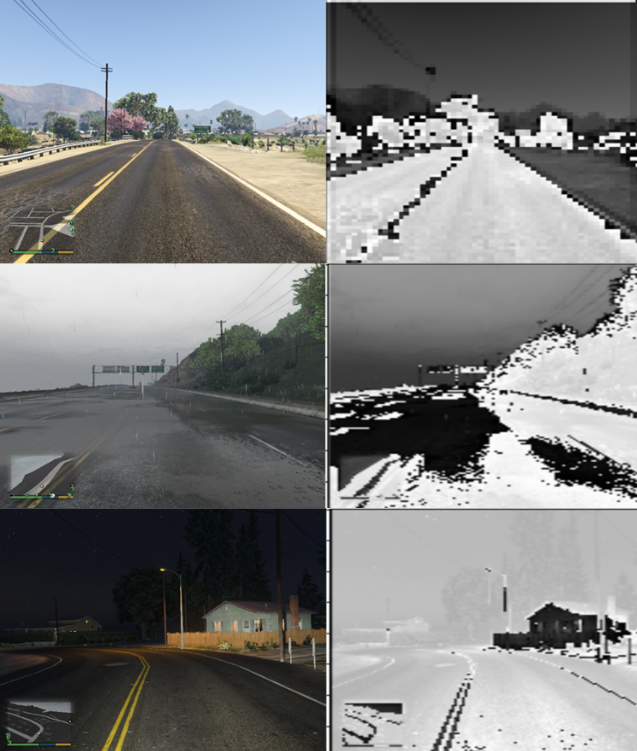}}
\caption{Input images and feature maps of the second convolutional layer in PilotNet}
\label{fig:feature_maps}
\end{figure}

These results can be explained by taking a closer look at the feature maps of the PilotNet model. Fig.~\ref{fig:feature_maps} shows the input images and the feature maps generated by the second convolutional layer of the original PilotNet model. The first row of images depicts driving during daytime in sunny weather. The PilotNet model accurately distinguishes the road from the surrounding environment and highlights the lane on the road, enabling the model to easily determine the steering wheel angle at that moment. The second row of images depicts the input image and feature map for driving in rainy conditions. The model's accuracy decreases in rainy weather because rain creates puddles on the road, causing the model to struggle to identify the road or lane markings clearly. The feature map shows the outline of puddles in the area where the model should have detected the lane and the road. The third row of images illustrates the scenario of driving at night with clear weather conditions. The model detects the lane only in the section of the road directly ahead of the vehicle. This is due to the illumination provided by the vehicle's headlights, which illuminate that portion of the road, while the surrounding areas remain dark. Such lighting conditions can present challenges, resulting in daytime driving being more accurate than nighttime driving. 

%% file: section/5_conclusion.tex
\section{Conclusion and future work} \label{conclusion}
The paper presents an autonomous driving system with a deep learning dual-model solution for steering and braking controls. Steering control is achieved using NVIDIA's PilotNet model for predicting steering wheel angles, while MobileNet SSD is utilized to detect vehicles ahead of an autonomous vehicle and to predict proximity information, which is subsequently used for braking control. The autonomous system's heavy computational demands come from utilizing two convolutional models, where the PilotNet model is not appropriate for deployment on smaller devices, leading to substantial inference latency. This significant latency can lead to lack of proper synchronization within the system, which increases the risk of crashes. Consequently, the PilotNet model was modified by replacing standard convolution layers with depthwise separable convolution layers, resulting in significant reduction of model parameters, size and inference latency. Additionally, bottleneck layers were introduced between the depthwise separable convolutional layers to reduce computational complexity, enabling faster training and inference without compromising model performance. Autonomous driving simulation and model evaluation was conducted using the Grand Theft Auto V video game. Both the original and modified PilotNet models were trained with the data containing diverse weather and lighting conditions, and road types. Evaluation of autonomous driving systems, with one utilizing the original PilotNet and the other employing the modified model version, was conducted across various scene conditions, revealing that the modified PilotNet model demonstrates comparable accuracy with decreased latency compared to the original model, while also being significantly smaller in size and thus appropriate to run on resource-constrained devices.\par

The future work will focus on enhancing the autonomous driving system in the following ways: The PilotNet model can be used to predict not only steering angle of the steering wheel but also throttle controls, while MobileNet SSD can be extended to detect various obstacles, such as pedestrians, traffic lights, and road signs. Another approach that could significantly enhance autonomous driving systems involves conducting neural architecture search on a modified PilotNet model. This process would try to identify the optimal architecture that maintains high accuracy, minimal inference latency, and a low number of parameters, while maintaining strong performance during evaluation in the simulator. The autonomous driving system can be evaluated in an alternative simulator with greater control over driving scenarios. Additionally, more data featuring driving in rainy conditions, at night, and on city roads can be incorporated in the dataset to increase PilotNet's performance in such scenarios.

%% file: paper.bbl
\begin{thebibliography}{10}
\providecommand{\url}[1]{#1}
\csname url@samestyle\endcsname
\providecommand{\newblock}{\relax}
\providecommand{\bibinfo}[2]{#2}
\providecommand{\BIBentrySTDinterwordspacing}{\spaceskip=0pt\relax}
\providecommand{\BIBentryALTinterwordstretchfactor}{4}
\providecommand{\BIBentryALTinterwordspacing}{\spaceskip=\fontdimen2\font plus
\BIBentryALTinterwordstretchfactor\fontdimen3\font minus \fontdimen4\font\relax}
\providecommand{\BIBforeignlanguage}[2]{{%
\expandafter\ifx\csname l@#1\endcsname\relax
\typeout{** WARNING: IEEEtran.bst: No hyphenation pattern has been}%
\typeout{** loaded for the language `#1'. Using the pattern for}%
\typeout{** the default language instead.}%
\else
\language=\csname l@#1\endcsname
\fi
#2}}
\providecommand{\BIBdecl}{\relax}
\BIBdecl

\bibitem{autonomous_driving_handbook}
R.~Matthaei, A.~Reschka, J.~Rieken, F.~Dierkes, S.~Ulbrich, T.~Winkle, and M.~Maurer, \emph{Handbook of Driver Assistance Systems: Basic Information, Components and Systems for Active Safety and Comfort}.\hskip 1em plus 0.5em minus 0.4em\relax Springer International Publishing, 01 2016, ch. Autonomous Driving, pp. 1519--1556.

\bibitem{autonomous_vehicles_review}
W.~Schwarting, J.~Alonso-Mora, and D.~Rus, ``Planning and decision-making for autonomous vehicles,'' \emph{Annual Review of Control, Robotics, and Autonomous Systems}, vol.~1, no.~1, 2018.

\bibitem{deep_learning_vehicle_control}
S.~Kuutti, R.~Bowden, Y.~Jin, P.~Barber, and S.~Fallah, ``A survey of deep learning applications to autonomous vehicle control,'' \emph{IEEE Transactions on Intelligent Transportation Systems}, vol.~22, no.~2, pp. 712--733, 2021.

\bibitem{deep_learning_survey}
J.~Ni, Y.~Chen, Y.~Chen, J.~Zhu, D.~Ali, and W.~Cao, ``A survey on theories and applications for self-driving cars based on deep learning methods,'' \emph{Applied Sciences}, vol.~10, no.~8, 2020.

\bibitem{autonomous_driving_survey}
E.~Yurtsever, J.~Lambert, A.~Carballo, and K.~Takeda, ``A survey of autonomous driving: Common practices and emerging technologies,'' \emph{IEEE Access}, vol.~8, 2020.

\bibitem{pilotnet}
M.~Bojarski, D.~W. del Testa, D.~Dworakowski, B.~Firner, B.~Flepp, P.~Goyal, L.~D. Jackel, M.~Monfort, U.~Muller, J.~Zhang, X.~Zhang, J.~Zhao, and K.~Zieba, ``End to end learning for self-driving cars,'' \emph{ArXiv}, vol. abs/1604.07316, 2016.

\bibitem{deeppicarmicro}
M.~Bechtel, Q.~Weng, and H.~Yun, ``{DeepPicarMicro: Applying TinyML to Autonomous Cyber Physical Systems},'' in \emph{2022 IEEE 28th International Conference on Embedded and Real-Time Computing Systems and Applications (RTCSA)}, 2022, pp. 120--127.

\bibitem{gta_vgg_lstm}
G.~A.~M. Novello, H.~Y. Yamamoto, and E.~L.~L. Cabral, ``An end-to-end approach to autonomous vehicle control using deep learning,'' in \emph{Brazilian Journal of Applied Computing 13 (3)}.\hskip 1em plus 0.5em minus 0.4em\relax Revista Brasileira de Computação Aplicada, 2021.

\bibitem{gta_vgg_pilotnet}
S.~R. Jaladi, Z.~Chen, N.~R. Malayanur, R.~M. Macherla, and B.~Li, ``End-to-end training and testing gamification framework to learn human highway driving,'' in \emph{2022 IEEE 25th International Conference on Intelligent Transportation Systems (ITSC)}.\hskip 1em plus 0.5em minus 0.4em\relax Institute of Electrical and Electronics Engineers (IEEE), 2022, pp. 4296--4301.

\bibitem{gta_alexnet}
M.~Martinez, C.~Sitawarin, K.~Finch, L.~Meincke, A.~Yablonski, and A.~Kornhauser, ``Beyond grand theft auto v for training, testing and enhancing deep learning in self driving cars,'' 2017.

\end{thebibliography}
